\documentclass{article}
\usepackage{ijcai20}

\usepackage{times}
\usepackage{soul}
\usepackage{url}
\usepackage[hidelinks]{hyperref}
\usepackage[utf8]{inputenc}
\usepackage[small]{caption}
\usepackage{graphicx}
\usepackage{amsmath}
\usepackage{amsthm}
\usepackage{booktabs}
\usepackage{algorithm}
\usepackage{algorithmic}
\newtheorem{mytheorem}{Theorem}

\newcommand{\mymin}{\mbox{\rm min}}

\newcommand{\myOmit}[1]{}

\title{Fair Division: The Computer Scientist's Perspective}

\author{
 Toby Walsh
    \affiliations
 UNSW Sydney and Data61
    \emails
    tw@cse.unsw.edu.au
}

\begin{document}

\maketitle

\begin{abstract}
I survey recent progress on a classic and challenging problem in
social choice: the fair division of indivisible
items. I discuss how a computational perspective has 
provided interesting insights into and understanding of
how to divide items fairly and efficiently. This has 
involved bringing to bear tools such as those used in 
knowledge representation, computational
complexity, approximation methods, game theory, online analysis
and communication complexity. 
\end{abstract}

\section{Introduction}

Fair division is an important problem facing society today
as increasing economical, environmental, and other pressures
require us to try to do more with limited resources. 
Consider allocating donated
organs to patients, pupils to schools, 
doctors to hospitals, 
donated food to charities, 
or water quotas to farmers. 
How do we do this both fairly and efficiently? 

Fair division is a problem with a long history. 
The ``divide and choose'' mechanism for fair division, for example, 
is mentioned in the Book of Genesis when Abraham and Lot 
use it to divide the land of Canaan. However, 
the theoretical study of fair division dates back only to the end of 
the Second World War when it attracted the
attention of mathematicians such as Hugo Steinhaus, Bronisław Knaster and
Stefan Banach \cite{steinhaus}. It later attracted
the attention of economists like Steven Brams and Vincent Crawford
(e.g. \cite{cakecut}). 

Much more recently, the fair division problem 
has attracted the attention of computer scientists, in particular
those working within artificial intelligence. 
We argue here that this has led to a range of computational
questions to be asked, and computational tools to be applied, that have
advanced both our understanding of and ability to find fair and 
efficient allocations. I focus this survey on the fair 
division of indivisible goods, but I could have just as easily
chosen the fair division of divisible goods. An emerging
trend, which I discuss briefly in the conclusion, is to the
fair division of indivisible bads (aka ``chores'') and of mixed manna
(goods and bads). 

\section{Formal model}

I briefly give some notation and concepts that will be 
used in the rest of this survey. 
We have $n$ agents and $m$ indivisible items, $a_1$ to $a_m$. 
For each agent $i\in [1,n]$ and $j \in [1,m]$, the agent $i$ has utility
$u_i(a_j) \geq 0$ for the item $a_k$. 
We will mostly limit ourselves to additive utilities where
the utility to agent $i$ for a set of items $B$ is $u_i(B) = \sum_{a \in B}
u_i(a)$. 
One special case of additive utilities is Borda scores
where the $j$th most preferred item for agent $i$ has utility $m-j+1$. 
Another special case of additive utilities is lexicographical scores
where the $j$th most preferred item for agent $i$ has utility $2^{m-j}$. 
We will also mostly consider fair division problems
involving just goods where items have non-negative
utility. This contrasts to bads where all agents have negative utility
for the items.
An allocation is a partition of the $m$ items into $n$ distinguished sets, $A_1$ to
$A_n$
where $A_i$ represents the items allocated to agent $i$ 
and $A_i \cap A_j = \emptyset$ for $i \neq j$. 
We consider a number of welfare notions for an allocation
$A_1$ to $A_n$: the utilitarian welfare which
is $\sum_{i=1}^n u_i(A_i)$, 
the egalitarian welfare which is $\mymin_{i \in [1,n]} u_i(A_i)$,
and the Nash welfare which is 
$\prod_{i=1}^n u_i(A_i)$.

A fundamental fairness notion in fair division is envy-freeness.
The ``divide and choose'' mechanism mentioned earlier
will return allocations that are envy-free assuming
risk averse agents. An allocation $A_1$ to $A_n$ is envy-free (EF) iff 
for any pair of agents $i,j \in [1,n]$ we have
$u_i(A_i) \geq u_i(A_j)$. Note that an envy-free
allocation may not exist. Consider two agents 
that both have positive non-zero utility for a single good.
A relaxation of envy-freeness for indivisible goods that can always be 
achieved is envy-freeness up to one good. 
An allocation $A_1$ to $A_n$ is envy-free up to one good (EF1) iff 
for any pair of exists $i,j \in [1,n]$ there exists a good $o_k$ with 
$u_i(A_i) \geq u_i(A_j\setminus\lbrace o_k\rbrace)$.

Another fundamental notion in fair division is Pareto 
efficiency. We don't want an allocation which all agents
think can be improved upon or at least not made worse. 
We say that an allocation $A_1$ to $A_n$ is Pareto dominated
iff there is another allocation $B_1$ to $B_n$ with
$u_i(B_i) \geq u_i(A_i)$ for all agents $i \in [1,n]$ and
for one agent $j \in [1,n]$ we have $u_j(B_j) > u_j(A_j)$. 
An allocation is Pareto efficient iff there is no
other allocation that Pareto dominates it. 

We consider mechanisms that allocate
items to agents. For instance, with the 
sequential allocation mechanism, agents take turns
to pick an item according to a picking order. 
This can implement mechanisms
like round robin (aka ``draft'') where
agents pick one item each in a fixed 
order, and this fixed order is then repeated until no
more items are left. 
Another interesting mechanism is the probabilistic 
serial mechanism \cite{probserial}. 
Agents simultaneously ``eat'' their most preferred remaining item
at some uniform rate. In general, this results in a randomized
allocation, which can be implemented as a probability distribution
over discrete allocations. An important property of 
mechanisms is whether they are strategy proof. Do agents
have any incentive to behave strategically? 
We say that a mechanism is strategy proof
iff it is in the best interests of an agent 
to act sincerely (e.g. in the case of the
sequential allocation mechanism, to pick
their most preferred item at every turn). 

\section{Knowledge representation questions}

A starting point for much research in AI/CS are 
questions of representation and reasoning. How, for instance, do 
represent a problem {\em compactly} so that we can reason
about it {\em efficiently}? In fair 
division, such questions abound. For example, 
how do we compactly represent an agent's preferences
over the power set of different possible subsets of items 
that might be allocated so that we can 
determine tractably an efficient and 
fair outcome? This power set of items creates an immediate
problem. We need to represent (and elicit) an agent's
preferences over an exponential number of possible
subsets of items. Often then a domain restriction
is supposed like dichotomous preferences or 
additive utilities to ensure preference can
be expressed compactly. However, this alone may not 
be enough to ensure reasoning is efficient. 

For example, Bouveret and 
Lang \shortcite{bljair2008} consider simple 
dichotomous preferences in which a subset of items is 
acceptable or it isn't and this is specified by (models of)
a propositional logical formula. Even in this 
simple and compact setting, deciding if an efficient
and envy-free allocation exists is intractable in general. 
Disappointingly, this problems remains intractable 
under strong domain restrictions such as identical
dichotomous preferences with just 
2 agents. 

\begin{mytheorem}[Proposition 6 and 7 in \cite{bljair2008}]
Deciding the existence of an efficient and envy-free allocation
with general dichotomous preferences specified by models of a propositional formular
is $\Sigma_2^p$-complete in general. The problem remains NP-hard 
with just 2 agents and identical, monotonic and 
dichotomous preferences. 
\end{mytheorem}

Another common domain restriction is to additive
utilities (e.g. \cite{bevia1998,brams2003,chevaleyre2008,keijzer2009}). 
Additive utilities offer a
compromise between simplicity and expressivity. 
Preferences are now compact to represent and elicit. 
However, we can no longer represent certain settings such as those
involving diminishing returns (e.g. a second bicycle is of
less utility to me than the first bicycle) or
complementarities (e.g. the cricket bat
is only of value to me with the cricket ball). 
Again, reasoning about even such a compact representation may 
still be challenging. For instance, 
finding an envy-free allocation with additive
utilities is NP-hard even under relatively 
specialized conditions such as additive 0/1 utilities
\cite{agmwaaamas14,agmwaij2015}.

\section{Computational questions}

One response to the sort of intractability results discussed in the last
section is to consider simple polynomial time mechanisms for
allocating items. 
Such mechanisms might return a ``good'' allocation, even if 
there is no guarantee of exact optimality. Such mechanisms can
also address issues like trust (when they are decentralized)
and manipulation (when they are strategy proof). 
Consider again the simple sequential allocation mechanism.
Variants of this mechanism are used in many real world settings such as assigning
students to courses at Harvard Business School \cite{budish2011}. 
In the sequential allocation mechanism, 
agents simply take turns to pick their most preferred
item. Whilst simple, sequential allocation has some nice theoretical properties. 
For example, if the mechanism is operated in a round robin
fashion then the allocation returned
is guaranteed to be envy-free up to one good \cite{budish2011}.

Such mechanisms for fair division invite 
a range of computational questions around their application. 
Consider 
again the sequential allocation mechanism. This is not
strategy proof in general. It may not always be in your best
interest sometimes to pick sincerely an item that is your most preferred
remaining item.
What then is the computational complexity
of computing optimal play (e.g. the subgame perfect Nash
equilibrium of the repeated sequential allocation game)? 
This problem remained open for a long time. 
Brams and Straffin
\shortcite{bsams79} first identified this 
computational question over forty years
ago: 
\begin{quote}
{\em ``no algorithm is known which will produce optimal
[strategic] play more efficiently than by checking many
branches of the game tree}
\end{quote}
It took more than three decades before 
computer scientists proved  that this computational problem was intractable 
in general even under restrictions on preferences of the agents. 

\begin{mytheorem}[Theorem 3 in \cite{knwxaaai13}]
With an unbounded
number of agents and additive Borda utilities, the
subgame perfect Nash equilibrium for the sequential allocation
game is unique and is PSPACE-hard to compute.
\end{mytheorem}

With just two agents, optimal play can be computed in linear
time (Theorem 1 in \cite{knwxaaai13}). 
Unfortunately, this optimal play may give rather poor outcomes.
The agent not making the final pick can play
strategically earlier in the game to ensure 
that their opponent taking the last (and thus only remaining) item
has their least preferred
item. Their opponent won't pick this item given any other choice,
but strategic play will ensure that this is their final choice. 
The agent not taking the penultimate pick can play
strategically earlier in the game to ensure 
that their opponent choosing an item in this penultimate pick
has their least preferred item, discounting the item that
will be picked in the final pick. And so on
for earlier picks. As a result, optimal strategic
play results in agents getting some of their 
least preferred items. 

With three or a larger but bounded number of 
agents, the computational complexity of computing 
optimal play remains stubbornly 
open (Challenge 5 in
\cite{waaai16}). 
It was recently shown that an agent's best response can be computed
in polynomial time for any bounded number 
of agents using dynamic programming \cite{xlaaai2020}. This, however,
leaves open the fixed point. 

\section{Calculation questions}

Alongside such computational questions, there are
many natural questions about calculation. 
For example, 
how do we calculate the welfare that agent can expect to have from applying 
a given mechanism? What is the probability that 
an allocation returned by a given mechanism is envy free?
What is the least utility an agent can get?
How do you calculate the best picking order?

Consider again the sequential allocation mechanism,
and a setting where the preferences of agents are fully independent 
(i.e. all possible preference orders are equiprobable and 
independent of each other). 
Bouveret and Lang \shortcite{indivisible} 
consider how to calculate the 
expected social welfare of allocations returned by applying the
sequential allocation mechanism in this setting. 
They conjecture that calculating the expected welfare is NP-hard
for a range of additive utilities
such as Borda and lexicographic scores given
the super-exponential number of possible 
preference profiles that need to be considered. Surprisingly, 
this conjecture turned out to be false.
Calculating the expected egalitarian, utilitarian or Nash
welfare takes only low-polynomial
time for any type of additive utilities (Lemmas 1 and 2 
in \cite{knwijcai13}). 

A related question is calculating the 
picking order that maximizes the expected 
social welfare. For small numbers of agents and items, 
Bouveret and Lang \shortcite{indivisible} 
calculate the optimal picking orders to
maximize the egalitarian or utilitarian
welfare supposing Borda utilities. They conjecture that calculating the optimal
picking order is NP-hard in general. For two agents and Borda utilities, 
it was subsequently shown that the optimal picking
order simply alternates between the two agents.

\begin{mytheorem}[Theorem 1 in \cite{knwijcai13}]
The expected utilitarian social welfare of the allocation
returned by the sequential allocation mechanism 
is maximized by the strict alternating picking order for two agents supposing
Borda utilities and the full independence assumption.
\end{mytheorem}

This result holds even if the two agents act strategically \cite{knwijcai13}.
For three (or more) agents, the complexity of computing the optimal
picking order remains open. 
For $k$-approval utilities (where
each agent has $k$ items with utility 1 and the rest with zero) 
it is not hard to see that
the strict alternating picking order may not be optimal even with just
two agents. Kalinowski {\it et al.} \shortcite{knwijcai13} 
conjecture that 
the strict alternating picking order is optimal for all convex scoring
functions (which includes both Borda and lexicographical scores). 

\section{Parameterized complexity questions}

Another tool in the computer scientist's toolkit in the face of 
intractability is parameterized complexity. Can we identify
parameters under which the computational problem becomes
polynomial? 
Consider again computing 
the subgame perfect Nash equilibrium for
the sequential allocation game. Recall that this is PSPACE-hard
in general.  Strategic behaviour is only
worthwhile when agents have different utilities
for an item. For example, if you value an item that I don’t, you may strategically
delay choosing it since it might still be available in a later
round. We say that an item is multi-valued if two agents
assign it different utilities, otherwise we say that the item
is single valued.
Kalinowski {\it et al.} \shortcite{knwxaaai13}
prove that, for any number of agents, we can compute
a subgame perfect Nash equilibrium in $O(k!m^{k+1})$ time
where $k$ is the number of multi-valued items, and $m$ is the
number of single valued items. Thus, when the number of 
multi-value items $k$ is bounded, the problem is polynomial. 

As a second and more recent example, 
Bredereck {\it et al.} \shortcite{bkknec2019}
provide fixed-parameter tractability results for a broad set of problems
concerned with the envy-free and efficient fair division of
indivisible items. 
For instance,
they show that computing an envy-free and efficient allocation of indivisible goods
is fixed-parameter tractable for the combined parameter of the number of agents together with the
maximum value of the utility of an item supposing 
additive utilities (Theorem 1 in \cite{bkknec2019}). 
In principle, this implies polynomial time envy-free and efficient allocations
supposing we have few agents and low maximum utility values.  
These results exploit a famous theorem of Lenstra concerning (the fixed-parameter tractability of)
Integer Linear Programs (ILPs) with bounded dimension. It remains
to be shown if the constructed ILPs are actually easy to solve in
practice.  
As the authors remark,
``algorithms designed for specific problems usually outperform this
approach [ILPs of bounded dimension] both in theory and practice''. 

\section{Game theoretic questions}

Not surprisingly, game theory provides an important
perspective for fair division. 
We have already mentioned some computational questions around
computing strategic actions. Recall, for instance, that
the subgame perfect Nash equilibrium for the sequential allocation
game is PSPACE-hard to computer in general, but linear time when
limited to just two agents. More generally, our understanding of fair division 
has advanced by asking questions from 
the interface between computer science and 
game theory. 

Consider, for instance, the probabilistic serial
mechanism \cite{probserial}. This is envy free and Pareto efficient
ex ante but is not strategy proof. The
probablistic serial mechanism is not strategy proof as an agent may not
``eat'' their most preferred remaining item if other agents do not 
value this item highly, in deferrence to eating
earlier a less preferred item that might otherwise be 
eaten by other agents. 
Although the probablistic serial mechanism is manipulable, finding an
optimal manipulation is in general computationally intractable even if
an agent has complete knowledge about the preferences of other
agents. Computational complexity
may in this case be a welcome barrier against agents behaving
strategically. 

\begin{mytheorem}[Theorems 2 and 3 in  \cite{agmmnwaamas2015}]
Computing a best response to the probablistic serial mechanism game
in order to maximize an agent's expected utility
is NP-hard in general. However, it takes linear time for just two agents. 
\end{mytheorem}

More generally, a pure Nash equilibrium is guaranteed to exist. 
Not surprisingly, given the complexity of best response, 
computing such an equilibrium is NP-hard. Indeed, even verifying whether 
the preferences reported by agents constitute a pure 
Nash equilibrium is coNP-complete (Theorem 3 in \cite{agmmnwijcai2015}]). 
On the other hand, with just two agents, a pure Nash equilibrium can be computed
in linear time (Theorem 5 in \cite{agmmnwijcai2015}]). Pleasingly, 
this results in the same outcome as sincere reporting. Thus, 
whilst strategic behaviours are easy to compute in the case
of just two agents, they do not change the final outcome. 

\section{Online questions}

Traditional models of fair division usually suppose 
all information about the problem is present at the same time. However, computer 
scientists are used to dealing with online problems
where the problem may be revealed over time. 
Indeed, many real world fair division problems have
such a form with either the agents, or 
the resources to be allocated, or both not being fixed
and potentially changing over time
(e.g. \cite{wki2014,aagwijcai2015,waaai2015,leftbehind,mswijcai2017,mswaies18,gdagmmwijcai2019,procacciaonlinepast,awaaai2020}). 
Consider allocating deceased organs to patients, donated food to
charities, water rights to farmers, etc. 
We often cannot wait till all resources are available, 
preferences known or agents present before starting to allocate the
resources. For example, when a kidney is donated, it must be allocated
to a patient within a few hours. As a second example, when food items 
arrive at a food bank, they must be allocated to charities promptly. 
As a third example, we might have to start allocating water
rights to farmers before we know how much rain will come. 
Such online problems introduce a range
of technical and other questions. 

First, how do we adapt offline mechanisms to an online setting?
Consider again the sequential allocation mechanism. 
Agents take turns to pick their most preferred item. But
what if an agent is not present? Or if it is, their most
preferred item is not? 
Second, how do adjust normative properties to take 
account of the online nature of such fair division problems? 
For instance,  Aleksandrov and Walsh \shortcite{awpricai2019} 
relax the definition of strategy-proofness
to suppose past decisions are fixed while future decisions could 
still be strategic. This online form of strategy-proofness is less
onerous to achieve.
On the other hand, the online nature of such fair division
problems can make other properties harder to achieve. 
For example, in offline fair division, Pareto efficiency and envy-freeness
are possible to achieve simultaneously, e.g.\ the allocations returned
by the probabilistic serial mechanism are Pareto efficient and
envy-free ex ante \cite{probserial}. In online fair division, however, 
such a combination is impossible to achieve in general.

\begin{mytheorem}[Theorem 9 in \cite{awpricai2019}]
No online mechanism for fair division
is Pareto efficient and envy-free ex ante.
\end{mytheorem} 

The reason that efficiency and fairness are not 
simultaneously possible is that the future is 
uncertain. An online mechanism must commit now to 
an allocation for the current item which ensures an efficient but
possibly not fair future, or a fair but possibly not efficient
future. 

\section{Asymptotic questions}

Another tool in the computer scientist's toolbox is asymptotic
analysis. Even in the offline setting,
desirable fairness property such as envy-freeness cannot be
guaranteed. Consider
two agents and one indivisible item that both agents
value. However, we can identify online mechanisms that limit how envy
grows over time. For example, \cite{procacciaonlineenvy} consider
whether mechanisms can achieve {\em vanishing envy}.  That is,
after $m$ items have arrived in an online fashion, if the maximum amount
of envy an agent has for any other agent is $envy_m$, can we 
design mechanisms so that the ratio of $\frac{envy_m}{m}$
goes to zero as $m$ goes to infinity?
This is indeed possible and, in fact, easy to achieve. 
\cite{procacciaonlineenvy}  show that randomly allocating items 
gives such an envy ratio that vanishes in expectation against an adaptive 
adversary. Unfortunately,
it is not possible to do better than such a ``blind''
mechanism. Indeed, 
this random mechanism is asymptotically optimal up to logarithmic factors.

Asymptotic analysis can also identify the welfare cost on insisting
on a fair allocation as we let the number of items grow. 
For instance, Caragiannis et al.\ \shortcite{ckkkwine2009}
define the price of envy-freeness as the ratio between
the optimal utilitarian welfare and the maximum utilitarian welfare
of any envy-free allocation and prove that 
this grows as $O(n)$ for $n$ items normalized to lie
within $[0,1]$.  Note that this result is for 
problems in which an envy-free allocation exists. 

\begin{mytheorem}[Theorem 5 in \cite{ckkkwine2009}]
For $n$ agents, the price of envy-freeness is at least
$\frac{3n+7}{9} - O(\frac{1}{n})$. 
\end{mytheorem}

\section{Approximation questions}

When faced with an intractable problem, computer scientists will 
often look for approximate solutions that can be 
computed efficiently. This has proved popular in
fair division, starting with the work of 
Lipton {\it et al.} \shortcite{lmmsec2004}
and an attractive relaxation of envy-freeness: envy-freeness up to one good (EF1). 
As noted earlier, the sequential allocation mechanism run
in a round-robin fashion returns an allocation that is EF1.
Unfortunately, whilst the allocation it returns is EF1, it may not be Pareto
efficient. The mechanism is limited to allocations in which agents get
a similar number of items. It may, however, be mutually
beneficial to swap one item allocated to one agent 
for multiple items allocated to a second agent. 

Caragiannis et al.\ \shortcite{mnw} 
have proposed instead computing an allocation that
maximizes the Nash welfare. Theorem 3.2 
in \cite{mnw} proves that this
is both EF1 and Pareto efficient. 
They also prove it provides a good approximation to another popular
(yet 
sometimes unachievable) fairness property, the maximin share
guarantee. 
The maximin share is the least amount of utility an agent receives
if they divide items into bundles and they get the least valuable
bundle. The Nash welfare mechanism is implemented and in daily use in the
spliddit.org website.
This leads Caragiannis {\it et al.} \shortcite{mnw} to conclude:
\begin{quote}
{\em 
``\ldots the MNW [maximal Nash welfare] solution exhibits an elusive combination of fairness and efficiency properties,
and can be easily computed in practice. It provides the most practicable approach to date
— arguably, the ultimate solution — for the division of indivisible goods under additive
valuations\ldots''}
\end{quote}
This claim is perhaps a little strong since it is NP-hard
to compute an allocation with
optimal Nash welfare \cite{nguyen2014}. As to whether it is 
the ``ultimate'' solution, 
Barman, Murthy and Vaish \shortcite{barman2018} 
recently gave a pseudo-polynomial-time algorithm for computing an
allocation that is both  EF1 and Pareto efficient. In addition
this mechanism provides an $1.45$-approximation of the optimal Nash welfare.

An weaker approximation than EF1 is envy-freeness up to any good
(EFX) \cite{mnw}. This lies strictly between EF and EF1. An allocation that is EF
is EFX, one that is EFX is EF1, but not the reverse. 
Intriguingly, it is not known whether an EFX allocation 
always exists. There are, however, some special cases
where we know that EFX allocations always exist.
For instance, Amanatidis {\it et al,} \shortcite{efx} 
prove that an allocation that maximizes the Nash welfare
is always EFX as long as there are at most two possible values for the
goods. With three or more distinct values this implication no longer
holds. As a consequence, an EFX allocation always exists
with two valued goods. 

Other fairness properties can be approximated in a similar way.
For instance, another fairness property is proportionality 
(does an allocation exist in which each of the $n$ agents
get at least $\frac{1}{n}$th of their total utility?). 
With additive utilities, envy-freeness implies proportionality
(but not vice versa). Although it is a weaker
property, it is not weak enough to ensure that 
proportional allocations always exist. 
However, PROP1 allocations that are proportional up to one good
always exist \cite{cfsec2017}. 
Indeed, Barman and Krishnamurthy \shortcite{bkaaai2019}
have recently given a polynomial time mechanism for computing
an allocation that is both PROP1 and Parto efficient.

\section{Communication complexity questions}

Another tool in the computer scientist's toolkit is communication
complexity. 
Plaut and Roughgarden \shortcite{prsoda2019}
initiate the use of the tools of communication complexity
to study fair division problems with indivisible items. 
Their primary question is to determine whether players need to exchange
an exponential amount of information to compute a fair allocation, or
whether the problem can be solved using only polynomial communication
between the agents.
Interestingly there results show a significant
difference between two and more than two agents.
For more than two agents, exponential communication is needed
to achieve an envy-free or proportional allocation 
even when randomization is allowed and valuations
are submodular. This exponential continues to hold even when
we demand allocations which are within a constant
factor of envy-free or proportional. 
With two agents, there are more promising results.
For example, with submodular valuations, there exists a deterministic
protocol to achieve a proportional allocation with communication
that is polynomial in the number of items.

\section{Empirical questions}

The final tool in the computer scientist's toolkit we will consider
is empirical analysis. We can code up these mechanisms
for fair division and run them on real and synthetic data. 
There are many empirical questions that we can then address.
For instance, how hard
are fair and efficient allocations to find in practice? 
Even if it is not guaranteed,
does a mechanism return envy-free allocations
with high probability? 
Do we observe hard instances around some phase
boundary as in other combinatorial domains like 
satisfiability? 
 
Such empirical analysis has been driven in part by access to real world
data from websites such as 
spliddit.org \cite{spliddit}
and PrefLib.org \cite{preflib}. 
Computers also make it easy to generate
synthetic data on which to test fair 
division mechanisms. For instance, 
Dickerson {\it et al.} \shortcite{dgkpsaaai14}
demonstrate that there is a
sharp phase transition from non-existence to existence
of envy-free allocations, and that finding such allocations
is hardest at that transition. They also prove
asymptotically that, even
when the number of items is larger than the number
of agents by a linear fraction, envy-free allocations are
unlikely to exist under modest assumptions on
the distribution of instances. On the other hand, when the number of
items is larger by a logarithmic factor, envy-free allocations
exist with high probability again under 
modest assumptions. 

\section{Conclusions}

The computer scientist's toolbox of questions
and methods has enriched our understanding and ability to
fairly and efficiently divide resources between multiple agents. 
It has let us identify the computational complexity 
of finding fair and efficient allocations, 
and to seek alternatives such as approximations
to envy-freeness like EF1 and PROP1 that can be achieved
tractably when exact methods are intractable in general. 
It has also lead to the identification of parameters 
that give fixed parameter tractability. 
Other tools in the computer scientist's toolbox
like asymptotic analysis and communication
complexity have also proved useful in understanding
fair division mechanisms. 
But, as can be expected, applying these tools
and answering these computational questions has
led inevitably to new and important questions.

One area of recent attention is in the fair division of bads (sometimes
called ``chores''), and in the fair division of goods and bads (sometimes 
called ``mixed manna''). 
See, for instance, \cite{bogomolnaia2017,arswaaai2017,aziz2018ai3,aciwijcai2019,bogomolnaia2019}. 
The fair division of bads is not symmetric to the fair division of 
goods. In particular, giving no goods to an agent
is pessimal, whilst giving no bads to an agent is optimal. 
For such reasons, results for the fair division of goods do not
always map onto analogous results for the fair division of bads. 

As well as interesting questions about the fair division of bads and
of mixed manna, there are many other important open questions 
puzzling the research community. I shall end by mentioning just a few: 

{\em 

Do EFX allocations always exist, even with 3 agents and additive
utilities?
How do we compute them efficiently? 

Are there (computational) phase transitions in other properties besides
envy-freeness? If so, how do their locations compare? 

How do we compute optimal play efficiently
for the sequential allocation mechanism and 3 agents?

Are there other appealing approximations of fairness 
that can be achieved tractably (especially 
in combination with Pareto efficiency)? 

Can we apply other ideas from CS/AI to help us understand
and solve fair division problems (e.g. automated reasoning
methods like SAT for theorem discovery, or machine learning
methods for the design of fair division mechanisms)? 

Does a computational lens help us understand and solve
other problems in social choice? 
}

\bibliographystyle{named}
\bibliography{/Users/tw/Documents/biblio/a-z,/Users/tw/Documents/biblio/a-z2,/Users/tw/Documents/biblio/pub,/Users/tw/Documents/biblio/pub2,online,mixed}

\end{document}